\documentclass[letterpaper, 10 pt, conference]{ieeeconf}  

\IEEEoverridecommandlockouts                              

\usepackage{graphicx, float}
\usepackage{algorithm, algorithmic}
\usepackage{hyperref}

\usepackage{ulem}
\usepackage{graphicx}
\usepackage{subcaption}
\usepackage{cite}
\usepackage{mathtools}
\usepackage[nolist,printonlyused]{acronym}
\usepackage{cleveref}

\title{\LARGE \bf
	Robot in a China Shop: Using Reinforcement Learning for Location-Specific Navigation Behaviour
}

\author{Bian Xihan and Oscar Mendez and Simon Hadfield
}

\begin{document}
	
%
%
%



\begin{acronym}[PTAMM] 
	
	\acro{ai}[AI]{Artificial Intelligence}
	\acro{rl}[RL]{Reinforcement Learning}
	\acro{ros}[ROS]{Robot Operating System}
	\acro{slam}[SLAM]{Simultaneous Location and Mapping}
	\acro{merlin}[MERLIN]{Multi-environment Reinforcement-Learning in Navigation}
	\acro{gnn}[GNN]{Graph neural network}
	\acro{trpo}[TRPO]{Trust Region Policy Optimization}
	\acro{ppo}[PPO]{Policy Optimization Algorithm}
	\acro{ac}[AC]{Actor-Critic}
	\acro{a3c}[A3C]{Asynchronous Advantage Actor-Critic}
	\acro{pnn}[PNN]{Progressive Neural Network}
	\acro{sota}[SOTA]{state-of-the-art}
	\acro{rnn}[RNN]{recurrent neural networks}
	\acro{icra}[ICRA]{International Conference on Robotics and Automation}
	\acro{iros}[IROS]{International Conference on Intelligent Robots and Systems}
				
\end{acronym}

	
	\maketitle
	\thispagestyle{empty}
	\pagestyle{empty}

	\begin{abstract}
		
		Robots need to be able to work in multiple different environments. Even when performing similar tasks, different behaviour should be deployed to best fit the current environment. 
		In this paper, We propose a new approach to navigation, where it is treated as a multi-task learning problem. This enables the robot to learn to behave differently in visual navigation tasks for different environments while also learning shared expertise across environments. 
		We evaluated our approach in both simulated environments as well as real-world data. Our method allows our system to converge with a 26\% reduction in training time, while also increasing accuracy.

	\end{abstract}

	\section{Introduction}
	
	Potential applications of modern robotics are becoming wider as the technology evolves.
	We are giving robots more tasks in more locations and allowing them to face more difficult challenges. 
	Robotics has moved from vacuuming a room to delivering take-out food, from working the assembly line to managing an entire factory. 
	As their functionality grows, so does the variety of their potential work environments. 
	Currently, robots are designed to operate in a single class of work environment. 
	It is feasible to treat a house as a single environment(although this is a simplification). However, it is no longer possible when the environment expands to an entire city.
	This limitation of the environment should be addressed promptly.
	
	To relieve robots from the limitation of a singular environment, it is only natural to look at human behaviour for guidance. 
	When entering a different environment, we often behave differently, and this behaviour change is not caused by the task at hand, but rather by the environment itself. 
	When we walk into a china shop, we are slow and careful to avoid collisions as opposed to walking down an empty hallway which emphasises speed rather than precision. 
	Robots need this ability to understand the environment they are in and change their behaviour accordingly. 
	Additionally, we, as humans, compartmentalize our knowledge and memory so we can effectively work with only a small portion of them rather than everything all at once. 
	We use this idea as further motivation for our multi-task learning approach.
	
	\begin{figure}[thpb]
		\centering
		\includegraphics[scale=0.38]{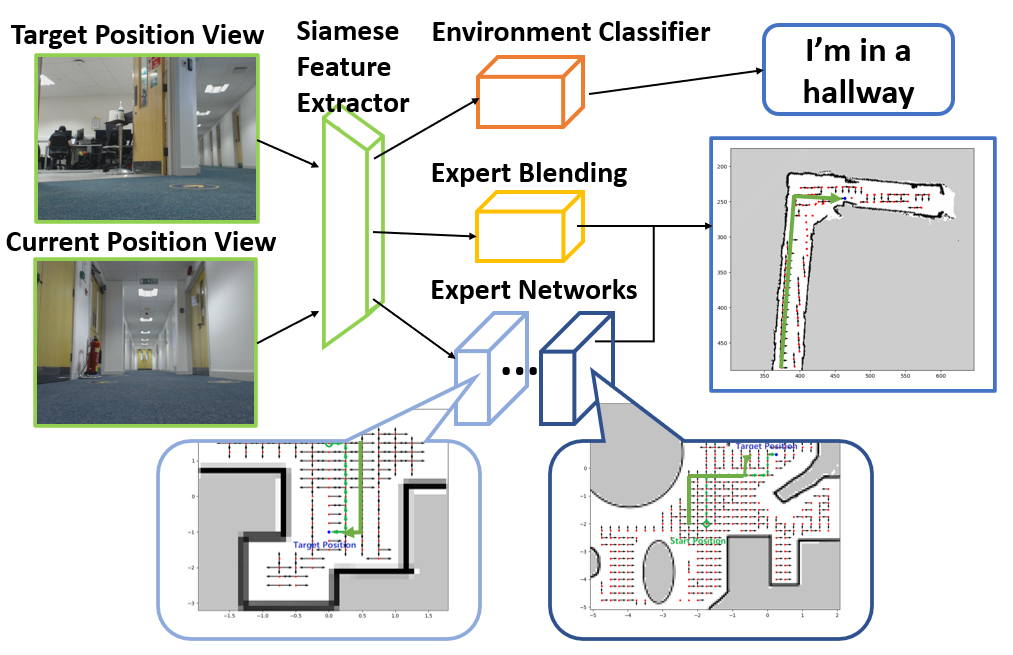}
		\caption{The target position view and current position view are given to the agent, the environment classifier encourages the network to distinguish between environments in earlier layers, while multiple expert networks' outputs are blended to produce the final policy}
		\label{fig0}
		\vspace{-5mm}
	\end{figure}
	
	In this paper, we define the combination of these abilities to be the problem of multi-environment navigation:
	A robot, operated by a single artificial intelligence model, should have the ability to recognize and navigate through multiple different types of environment. 
	To approach this problem, we use visual navigation, as visual information (compared to other sensory inputs) will be most effective in learning the difference between different environments. 
	In this paper, we focus on navigation from visual sensors, and propose a new network architecture: \acf{merlin}. 
	
	Visual navigation is a very active field of research, but most previous work focusses on improving navigation accuracy in a singular environment. 
	In this work, we provide a new architecture where visual navigation in multiple environments can be achieved.
	We utilize a Siamese network feature extractor and multiple expert networks with attentive gating, combined with a special classification branch to encourage the network in recognizing the difference between environments.
	We find our model outperforms the state-of-the-art visual navigation models.
	Additionally, by encouraging the classification of environments, we are also able to achieve better results in learning performance, accelerating learning in multi-environment navigation.
	
	In summary, we define a new problem of multi-environment navigation and propose a new model for approaching this problem. 
	The main contributions of the paper are as follows:
	\begin{enumerate}
		\item Define the problem of multi-environment navigation.
		\item Propose a new multi-task learning approach to visual navigation.
		\item Propose the inclusion of an intermediate environment classification loss to accelerate learning.
	\end{enumerate}

	\section{Related Work}
	
	\subsection{Visual Navigation}
	
	Visual Navigation relies on vision as the primary source of information for navigation, compared to other navigation methods which rely on distance sensors.
	In map-based approaches for visual navigation, visual input is used mainly for landmark tracking.
	The algorithms detect landmarks on the camera input and track them in the following frames to determine the position of the robot\cite{kabuka_position_1987,hashima_localization_1997}.
	Another approach is to let the robot explore the environment and instead of a map, the robot builds a feature representation model of the environment\cite{moravec_stanford_1983}.
	
	In more recent studies, mapless and \ac{ai}-based navigation is becoming more popular. 
	Mapless approaches (also known as visual odometry) \cite{nister2004visual}, include two main solutions: Optical-flow and Appearance-based matching.
	Optical-flow-based solutions estimate the motion of objects by tracking the motion of features throughout a sequence of visual input \cite{horn_determining_1981,lucas1981iterative}.
	Appearance-based matching solutions rely on prior knowledge of stored images of the environment \cite{kim_end--end_2018}.
	The robot will try to match the current view with the stored images to locate and navigate \cite{zhu_target-driven_2017,bai_toward_2017}.
	This idea of feature tracking to localization and navigation has been fundamental for \ac{ai}-based localization and navigation. 
	Notably, the work of Kendall in PoseNet \cite{kendall_posenet:_2015}, which uses a convolutional network for real-time camera relocalization where the model is trained on labelled images of the environments.
	
	We approached the problem of multi-environment navigation through target-driven navigation, which is a branch of appearance-based matching solutions. This approach employs the use of deep \ac{rl} which does not require supervised training for landmarks or features. 
	The work of Zhu et al. \cite{zhu_target-driven_2017} presents a \ac{sota} target-driven visual navigation solution.
	The model they proposed is an actor-critic model which has a policy function of both the goal and the current state as input.

	\subsection{Reinforcement Learning}
	
	
	Schulman's 2015 work in \ac{trpo} is a classic \ac{rl} algorithm that utilizes a different approach \cite{schulman2015trust}.
	Trust region methods use a model function to estimate the objective function, and by optimizing the model function, perform better actions.
	The size of the policy update is constrained to monotonically decrease to ensure convergence.
	Schulman's later work in the Proximal Policy Optimization Algorithms (\ac{ppo} and \ac{ppo}2) improved upon \ac{trpo} \cite{schulman2017proximal}. 
	These algorithms use a ``surrogate" objective function to determine the next action while interacting with the environment.
	It assumes that with similar state input, the agent should take similar action, which lowers the rate and necessity of re-sampling.
	Instead of constricting the model functions, \ac{ppo} applies a penalty to policies that differ from the objective function.
	
	Another \ac{rl} algorithm with a combined approach is the work of Mnih et al. in the \ac{ac} series of techniques \cite{mnih_asynchronous_2016}.
	These algorithms combine the Q-learning and policy gradient by creating a policy-based actor that chooses actions and a value-based critic which scores the actions.
	This allows the handling of both discrete and continuous problems, as well as updating more regularly for better learning efficiency. 
	In the more advanced version: \ac{a3c}, the algorithm allows asynchronously update to and get updated by the main policy by multiple agents, this advancement greatly increased computation and sampling efficiency, allows for faster training given the growing computing power. 
	The multi-agent support also enabled \ac{rl} in multi-task learning, allowing one agent to learn multiple tasks simultaneously.
	Due to its asynchronous nature, it is an ideal fit for multi-task learning by allowing agents to learn multiple tasks simultaneously, and update to a single main policy. 
	
	\subsection{Multi-Task Learning}

	In multi-task reinforcement learning, numerous works have shown the capability of a single agent to perform at an expert level in multiple Atari games using deep Q-learning \cite{mnih_human-level_2015, badia2020agent57}.
	In the \ac{rl} environment, the model makes no assumption of the relatedness of tasks, which enabled many different approaches such as policy representation for each task and regionalized policy clustering, \cite{kulkarni_hierarchical_2016} which employs a hierarchical Bayesian approach to model the distribution over Gaussian process temporal-difference value functions for each task, and \ac{a3c}-based deep reinforcement learning approach.
	The challenges these approaches all face are negative learning and scalability. 
	Negative learning refers to when agents `forget' previously learned knowledge while learning a new task. 
	The most popular solution to negative learning is through the use of a gating mechanism, where the network is only allowed to update part of itself during the training for each task. 
	This idea is first proposed by Rusu et al. in the work of \ac{pnn} \cite{rusu_progressive_2016}, where the network freezes itself and add new resources for the new task.
	In the training process of the new task, the network stops the updating of its current weights, then widens all the nodes to provide new resources for learning the new knowledge.
	This approach does provide a good solution to the problem of negative learning. However, it fails in scalability, due to  increasing learnable parameters and poor sampling efficiency with respect to the increasing complexity of tasks. 
	To improve sampling efficiency, the work of Andrychowicz et al.  \cite{andrychowicz_hindsight_2017} and Landolfi et al.  \cite{landolfi_model-based_2019} propose the use of a memory bank or model based approach during sampling. 
	For parameter size, a common solution to this is using model compression techniques for data efficiency, such as the work of Teh et al. in Distral \cite{teh_distral_2017}, where the network shares a distilled policy that captures the common behaviours across all tasks and allows workers to solve their own task while staying close to the main policy.
	
	In this study, we develop a multi-branch gated network somewhat similar to PNN\cite{rusu_progressive_2016} but with soft blending and lifelong training of all branches training simultaneously on all tasks, instead of sequential training and freezing weights. This framework makes no assumption on the relatedness of tasks and mitigated negative learning effect through attention based soft gating.

	\section{Methodology}
	
	\begin{figure*}[thpb]
		\centering
		\centering
		\includegraphics[scale=0.75]{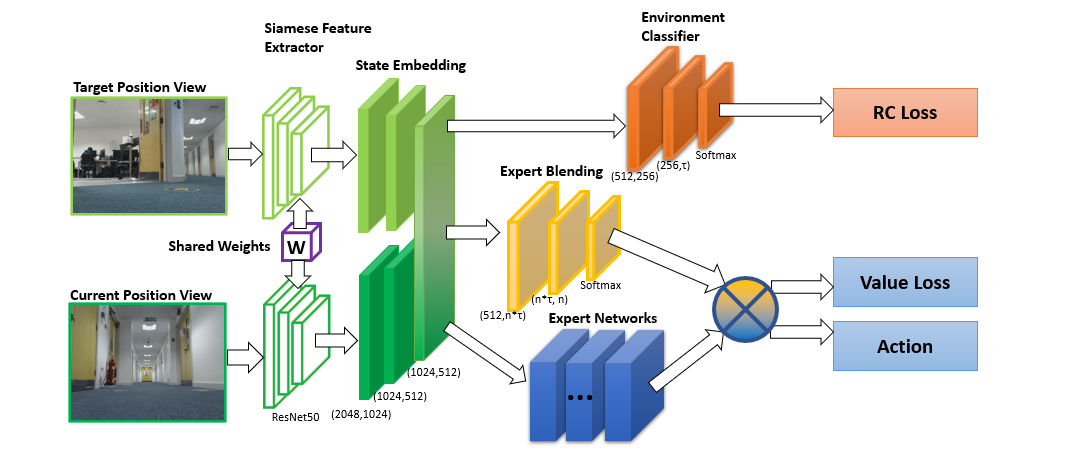}
		
		\caption{Our architecture can be divided into 4 major components: A Siamese feature extractor (green), $n$ sub-expert-networks (blue), attention network (yellow), and a environment classifier network (orange).}
		\label{fig:arch}
		\vspace{-5mm}
	\end{figure*}
	
	The objective of the agent is to be able to navigate in multiple types of environments. 
	The agent can navigate the environment through actions: step forward, step backward, turn right and turn left (90 degrees).
	Given a target, the agent will only be given its current observation $I_{O}$ and a view of the target $I_{t}$ . 
	A single agent should be capable of solving a set of similar navigation tasks placed in different environments by training a policy $ \pi (a_{t}|s_{t}, \tau) $ and value function $V(s_{t}, \tau)$ for each task, while maximizing the reward for each task.
	
	To tackle the specific problem of multi environment/task visual navigation, we propose a new architecture: \ac{merlin}. 
	Our architecture can be largely divided into 4 major components as shown in figure \ref{fig:arch}: A Siamese feature extractor, $n$ sub-networks, an attention network and an environment classifier network. In the following section, each of these components will be described in turn.
	
	\subsection{Siamese Feature Extractor and Joint State Embedding} 
	
	The network has 2 branches joined near the input to form a Siamese network. 
	The expected functionality of the Siamese feature extractor is to capture the input state through feature extractors as well as to identify  commonly useful information for all tasks and the feature characteristics to identify different tasks.

	%
	
	The Siamese feature extractor takes input in the form of a vector of the 2 images: the current agent observation  $I_{C}$ and the current target $I_{T}$. 
	Each image will be fed into a different branch of the Siamese feature extractor network. 
	Both the target state input and the current state input are first put through convolutional feature extractors $E_f$.
	The feature extractors share the same weights $W_{S}$ between branches, and the output of each branch will then go through a normalization function $N$ before concatenating into a state embedding.
	For the current state observation $O_{C}=(I_{T},I_{C})$ that contains the feature information from both input images, the features $F(s)$ is provided through the following equation:
	
	\begin{equation}
	\begin{multlined}
	F(s) = \{N(E_f(I_{T}|W_{S})),N(E_f(I_{C}|W_{S}))\}
	\label{atb:eq_0}
	\end{multlined}
	\end{equation}
	
	The Siamese feature extractors are updated by losses flowing through both the RC network branch and the sub-network/attentive network branch. This means the features are required to simultaneously be effective at solving the \ac{rl} navigation task, and capable of distinguishing different categories of the environment.
	
	\subsection{Task-Specific Expert Policy Sub-Networks}
	
	The $n$ sub-networks serve as the expert networks that learn the knowledge and skill to solve a specific task. 
	The number $n$ is determined by a range of factors including the number of tasks, the similarity between each task, the similarity between each environment, etc.
	In this work, the number $n$ equals the number of environments.
	This can be considered as an optimization between requirements and availability of resources for the agent.
	The optimizer will learn to ignore excess sub-networks when given more resources than required as shown in the work of Bram \cite{bram_attentive_2019}.
	
	The specific architecture of the sub-networks can be altered to fit the scenario. 
	It is possible to have a variety of different expert networks that would work better for different tasks or task settings combined within the same agent.
	As we are working on the subject of visual navigation, the sub-network architecture in this study is designed for visual information processing and navigation tasks:
	A softmax layer maps the last hidden layer of each network to an $A$ dimensional vector to produce action probabilities $A_{i}$ and a linear layer outputs the value function  $V_{i}$ for each expert network $i$ with a specific policy $\pi_{i}$:
	\begin{equation}
	A_i = softmax( \pi_i(E_i(F(s) | W_i) | W_A)
	\end{equation}
	\vspace{-5mm}
	\begin{equation}
	V_i = E_i(F(s) | W_i) \cdot W_v
	\end{equation}
	Where $E_i$ represents the \ac{rl} network used for the actor-critic branch.
	If the action space is different between tasks, the size of $A$ should be the largest action space size of all tasks. 
	
	\subsection{Attentive Task Allocation and Soft Blending Network}
	
	The attentive task allocation network first takes the output and recognizes the corresponding expertise required by different tasks. 
	While each expert sub-network produces a policy function, the attentive network assigns a distribution weight $W_{\tau}$ to these policy functions according to the estimated relevance of expertise.
	\begin{equation}
	W_{\tau} = softmax\left(\left\{Att(F(s) | W_{Att_{i}}) | i \in \{0..n\}\right\}\right)
	\end{equation}
	The softmax normalizes the weights $W_{Att_{i}}$ to sum to 1. 
	The final policy is then based on the dot product of the attention weights against the experts' action distributions: 
	\begin{equation}
	\pi(\alpha | s) = \sum_{i=0}^n W_{\tau_i} \cdot A_i
	\end{equation}
	where $W_{\tau_i}$ is the $W_\tau$ for the $i$th expert network.
	This policy will determine the action taken by the agent during each timestep. The attention weight $W_{\tau}$ is also used to compute a value function for reinforcement learning branches of the network. For the expert values $V_{\tau} = {V_{0}, V_{1} ... V_{n}}$, the combined value function is:
	\begin{equation}
	V_{rl}(s) = \sum_{i=0}^{n} W_{\tau_i} \cdot V_{i}(s)
	\vspace{-1mm}
	\end{equation}
	This value loss is used to update the expert networks, the attentive task allocation network, and the Siamese feature extractor, but not the RC networks.

	\subsection{Environment Classifier}
	
	Finally we propose an additional environment classifier, or Room Classifier (RC), network branch. This serves as a regularization that encourages the feature extractor layers to preserve the information that helps identify the tasks. 
	The RC network updates only itself and the feature extractor layers with a loss function depending only on the current state:
	\begin{equation}
	L_{rc} (s) = 1 - \mathcal{RC}(F(s) | W_{rc}) \cdot \mathcal{M_{\tau}}
	\vspace{-1mm}
	\end{equation}
	
	with $\mathcal{M_{\tau}}$ being a one-hot vector containing zeros everywhere, except the entry corresponding to the ground-truth task label which is 1.
	The RC loss function will be independent of the \ac{rl} loss function, updating only the RC network and the Siamese feature extractor. 
	
	In this work, the RC network is trained through supervised classification methods, it may be possible in the future to extend the capability of the RC network to recognize a new task that the agents have never seen using unsupervised techniques.

	\begin{table*}[!tb]
		\centering
		\resizebox{0.8\textwidth}{!}{%
			\begin{tabular}{|l|r|r|r|r|r|r|r|r|}
				\hline
				Sim Dataset  & Env1           & Env2           & Env3          & Env4          & Ep. Length     & Avg. reach goal & RC Accuracy & Converge Step \\ \hline
				\ac{merlin}     & 15.83          & \textbf{17.32} & \textbf{8.74} & \textbf{7.71} & \textbf{12.38} & \textbf{99.50}  & 99.61      & 23K           \\ \hline
				Expert1      & \textbf{13.14}          & 200.00         & 199.22        & 197.64        & 152.31         & 25.40           & 0.00       & 6K            \\ \hline
				Expert2      & 196.85         & 24.48          & 197.62        & 194.54        & 153.20         & 26.00           & 0.00       & 10K           \\ \hline
				Expert3      & 200.00         & 199.21         & 20.25         & 195.38        & 153.53         & 24.60           & 0.00       & 5K            \\ \hline
				Expert4      & 198.44         & 199.20         & 197.68        & 8.73          & 150.82         & 25.70           & 0.00       & 7K            \\ \hline
				Joint Expert & 15.53 & 21.32          & 11.49         & 8.96          & 14.30          & 99.30           & 0.00       & 31K           \\ \hline
			\end{tabular}%
		}
		\vspace{-1mm}
		\caption{Navigation task step count and success percentage in the simulated dataset}
		\label{table1}
		\vspace{-4mm}
	\end{table*}
	\begin{table*}[!tb]
		\centering
		\resizebox{0.8\textwidth}{!}{%
			\begin{tabular}{|l|r|r|r|r|r|r|r|}
				\hline
				Real Dataset & Env5           & Env6           & Env7           & Ep. Length     & Avg. reach goal & RC Accuracy     & Converge Step \\ \hline
				\ac{merlin}     & \textbf{10.82} & 40.27 & \textbf{25.04} & \textbf{25.35} & \textbf{94.60}  & 98.93 & 19K           \\ \hline
				Expert1      & 13.80          & 194.72         & 198.80         & 135.65         & 33.50           & 0.00           & 4K            \\ \hline
				Expert2      & 199.40         & 32.31          & 198.81         & 143.39         & 30.50           & 0.00           & 20K           \\ \hline
				Expert3      & 198.80         & 198.82         & 31.31          & 142.87         & 31.00           & 0.00           & 6K            \\ \hline
				Joint Expert & 18.05          & \textbf{21.94} & 38.86          & 26.28          & 94.10           & 0.00           & 25K           \\ \hline
			\end{tabular}%
		}
		\vspace{-1mm}
		\caption{Navigation task step count and success percentage in the real-world dataset}
		\label{table2}
		\vspace{-6mm}
	\end{table*}
	
	\section{Experiments and Results}
	
	To evaluate our approach, we experimented with our agent's ability to perform visual navigation tasks in both simulated and real environments. 
	The agent will be trained in multiple different themed environments simultaneously. 
	The \ac{rl} algorithm used in the experiments is a multi-thread \ac{a3c}. 
	The network backbone for the sub-networks are the same as the \ac{sota} model \cite{zhu_target-driven_2017}.
	The reward is inversely related to the length of the path taken by the agent to reach the target position: Negative reward accumulates with the path length $l$ at a rate of $p_{stp}$ per step. The agent also takes an additional penalty $p_{cr}$ for the number $n_{cr}$ when agent hitting obstacles.
	Each episode is limited to $t_{lm}$ timesteps to deter reward hacking and avoid the agent getting stuck. 
	Reaching the target position before this limit will result in a small positive reward $r_{ter}$.
	Exceeding this limit will result in ending the episode and a large negative reward $p_{ter}$. 
	The final reward $R$ is given by:
	\begin{equation}
	R = -p_{stp} * l - p_{cr}*n_{cr} + (r_{ter}|l \leq t_{lm}) - (p_{ter}|l>t_{lm})
	\vspace{-2mm}
	\end{equation}

	The agent will receive an image of the target position $I_T$ and the current position's view $I_C$.
	In simulation a small amount of random noise is added to the current position before sampling $I_C$ to ensure generalization.
	
	Our agent is implemented with Pytorch and trained on Nvidia Geforce GPU servers. 
	The evaluation is done in several experiments both in simulation and the real-world.

	\subsection{Experiments}
	We prepared both simulated and real-world environments for the experiments.
	To create the simulated environments, we used the 3D simulation environment AI2-THOR\cite{ai2thor}. 
	AI2-THOR is a 3D simulation program used for machine learning.
	The simulated environment consist of themed rooms such as living room, bathroom, kitchen, etc.
	We created the regularly sampled environments by allowing the agent to move forward and backwards as well as turning 90 degrees on a square grid.
	The agent will be dropped randomly into the environment and given a random target which is reachable from the starting position. 
	The square grid $g_{i} \in G$ has a grid size of $\alpha$ with $g_{i}$ being a sample from SO(2) (i.e. comprising of $x,y,\theta$)
	The target is described through the view of the agent $I_{T}$ at the target position $g_{T}=G(x_{T},y_{T},\theta_{T})$. 
	The current position $g_{C}=G(x_{C},y_{C},\theta_{C})$ of the agent is given through a view of the agent at its current position $I_{C}$.
	To ensure the generalization capability of the agent and simulate the navigation error of an actual robot, the current view is sampled randomly near each grid point.
	The random sample position is selected by adding Gaussian noise proportional to the grid size $\Delta g \sim \mathcal{N} (0,\,(\alpha*\rho)^{2})\, $ to the $x$ and $y$ coordinate of the original state position.
	Resulting in the view at $g_{sample}=G(x+\Delta x,y+\Delta y,\theta)$ as the actual input to the agent.
	
	The agent will then try to find the shortest path to reach the target position. 
	In our experiments, 4 environments are trained simultaneously, each using the same amount of computing resources, the average episode length and percentage of successful runs across all 4 environments are used to evaluate their performance.
	
	To train the agent in the real-world environments, we collected real-world data using a Turtlebot from various locations with different themes such as a hallway, living room, common room, etc. 
	As the robot already has drifting errors, there's no random sampling used in the real-world datasets.
	Using these images, we produced gridded real-world environments similar to the simulated environments which can be trained off-line.
	A total of 3 different real-world environments were used for training:
	A residential living room, a university common room, and an office corridor.
	They represent 3 types of spaces: a small enclosed space with few obstacles, a large open space with many obstacles, and a narrow enclosed space without obstacles.
	The agent is expected to behave differently in each of these types of space.
	
	In the last section of the experiments, we improve on this and demonstrate our model directly on a live robot.
	To achieve this, a Turtlebot is used to navigate in the previously trained environments.
	A video showing this experiment can be found at \href{https://youtu.be/ayjwNdCOkbw}{this link}.

	\begin{figure*}[h]
		\centering
		\parbox{\textwidth}{
			\begin{subfigure}{.35\textwidth}\hspace{0pt}
				\includegraphics[width=5.5cm, height=4cm]{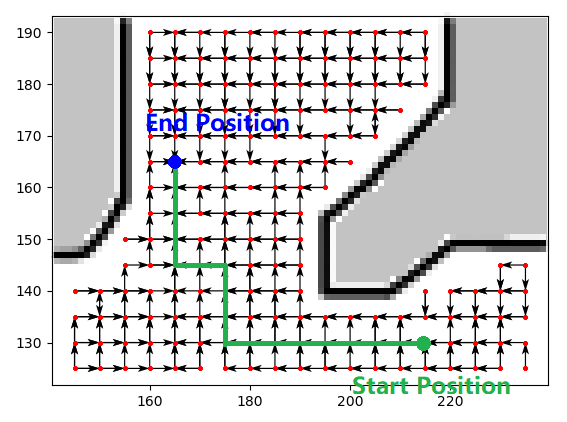}
				\vspace{-2mm}
				\caption{Env1: simulated kitchen}
				\label{exp2_1}
			\end{subfigure}\hspace{0pt}
			\begin{subfigure}{.35\textwidth}
				\includegraphics[width=5.5cm, height=4cm]{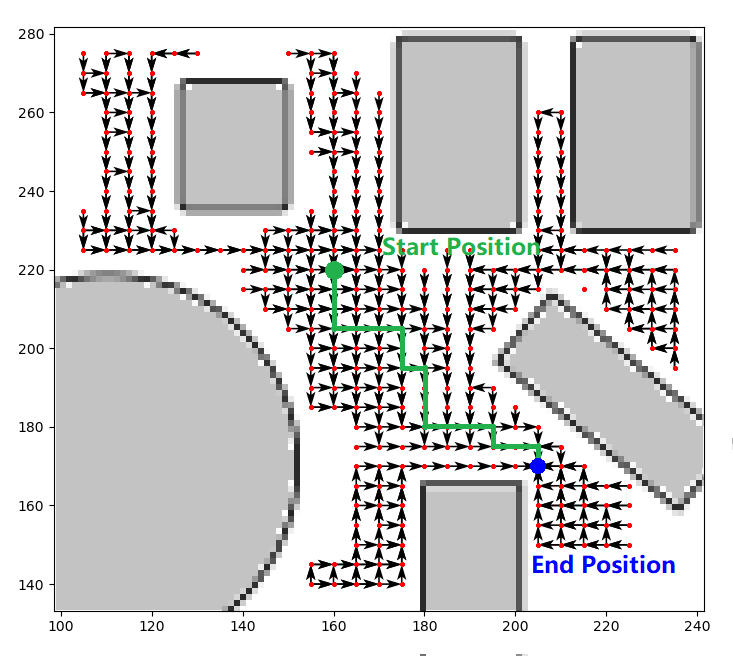}
				\vspace{-2mm}
				\caption{Env2: simulated living room}
				\label{exp2_2}
			\end{subfigure}\hspace{0pt}
			\begin{subfigure}{.35\textwidth}
				\includegraphics[width=5.5cm, height=4cm]{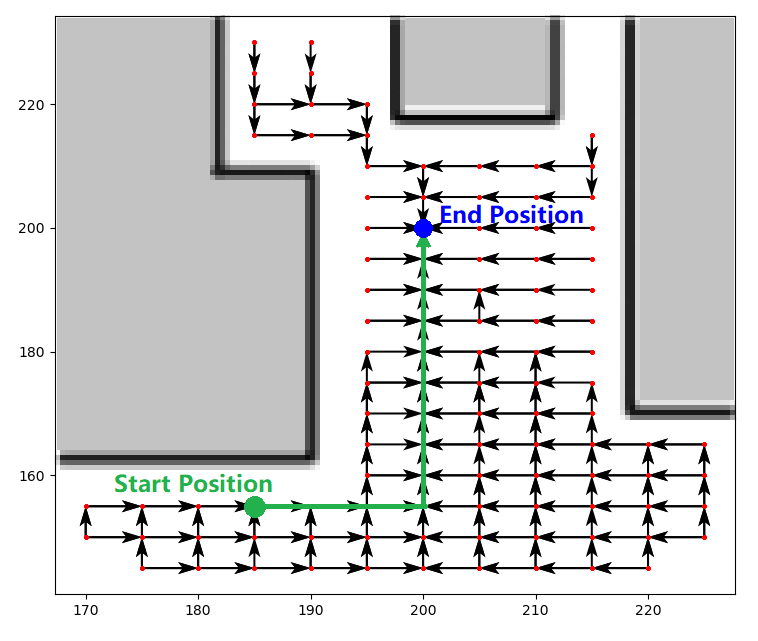}
				\vspace{-2mm}
				\caption{Env3: simulated bathroom}
				\label{exp2_3}
			\end{subfigure}\hspace{0pt}
			\begin{subfigure}{.35\textwidth}
				\includegraphics[width=5.5cm, height=4cm]{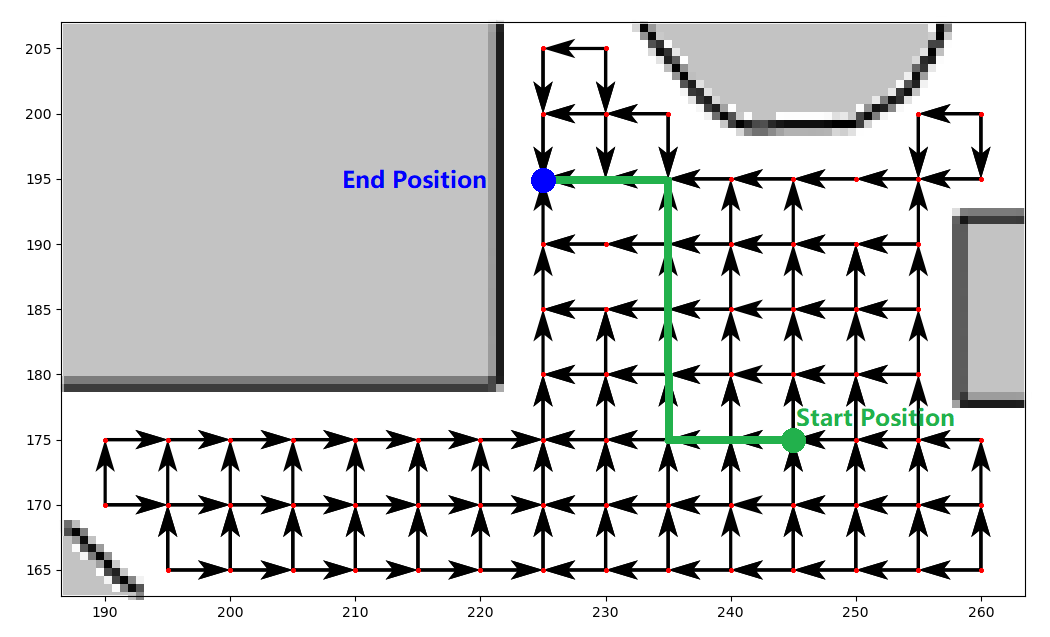}
				\vspace{-1mm}
				\caption{Env4: simulated tiny bathroom}
				\label{exp2_4}
			\end{subfigure}\hspace{0pt}
			\begin{subfigure}{.35\textwidth}
				\includegraphics[width=5.5cm, height=4cm]{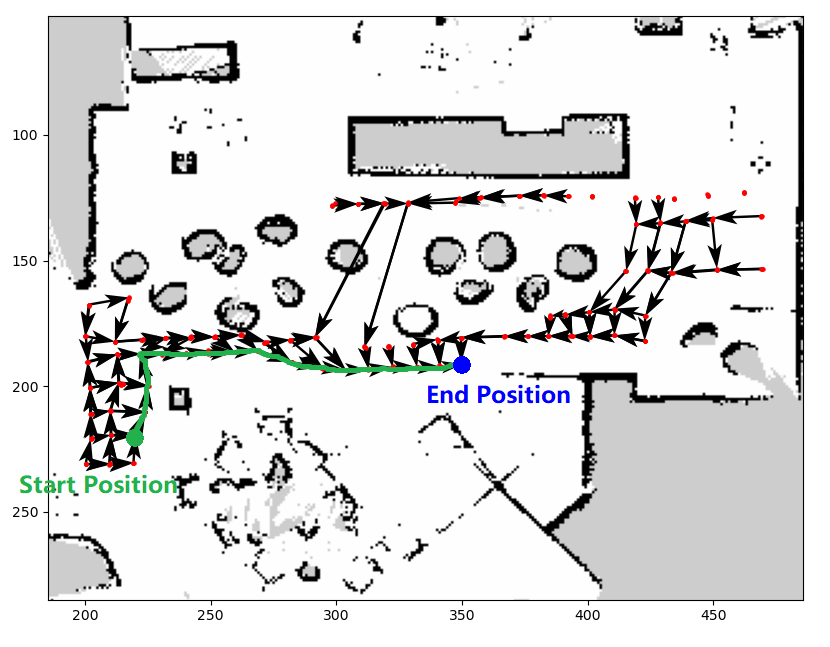}
				\vspace{-1mm}
				\caption{Env6: real common room}
				\label{exp2_6}
			\end{subfigure}\hspace{0pt}
			\begin{subfigure}{.35\textwidth}
				\includegraphics[width=5.5cm, height=4cm]{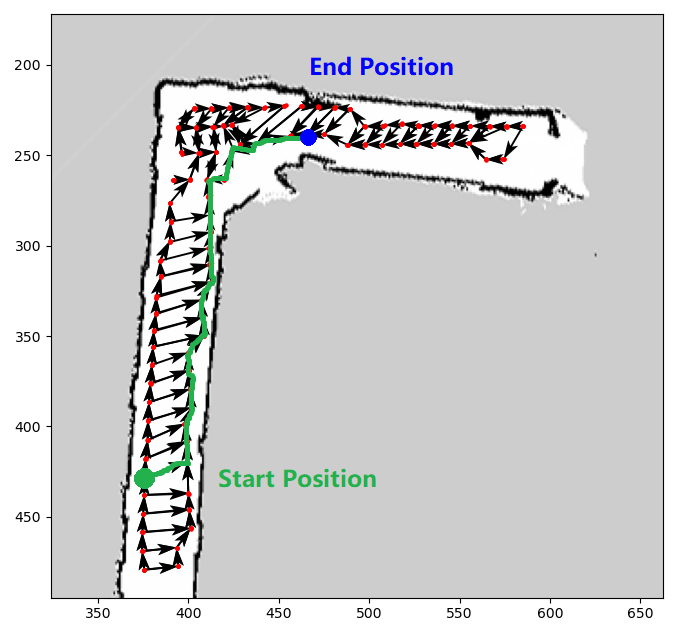}
				\vspace{-1mm}
				\caption{Env7: real corridor}
				\label{exp2_7}
			\end{subfigure}
		}
		
		\caption{Agent makes large strides before turning in open spaces, and avoiding walls when in narrow spaces}
		\label{fig3}
		\vspace{-3mm}
	\end{figure*}
	
	\subsection{Baseline comparison}
	
	We first compare our agent with the \ac{sota} target-driven navigation model \cite{zhu_target-driven_2017} and perform an ablation study of our technique. 
	The \ac{sota} model is trained in different environments both separately (referred to as Expert1 to Expert4) and concurrently (referred to as ``Joint Expert"). 
	All models are trained until they converge to over 99\% success rate, the average episode length and the number of time steps taken to converge are recorded.
	
	As shown in both Table \ref{table1} and \ref{table2}, the separately trained \ac{sota} models show an inability to perform navigation task in any environment other than the one it was last trained on. 
	The jointly trained baseline can complete the tasks in all different environments, but it requires a 30\% longer training time and has a lower performance than our proposed technique. 
	Additionally, our agent is able to out-perform the specialist experts in most of their corresponding environments.
	This suggests that there is a sharing of expertise between environments, which can take advantage of our expertise-blending approach.
	In the simulated kitchen, as shown in Figure \ref{exp2_2} and Table \ref{table1}, it is a larger environment with more grid points compared to other environments.
	The joint expert performs more poorly in this environment compared to its performance in the rest of the environments.
	A similar pattern can be observed in the real-world results as shown in Table \ref{table2}, the joint expert has a particularly low performance in the largest environment Env7.
	It appears that the improvements offered by the proposed approach scales with the number and the size of the environments.
	The \ac{merlin} model also has a 24\%-26\% faster converge speed.
	This indicates that when the number of learnable parameters increases, it is possible that our approach will still be able to converge on more difficult tasks when the joint expert cannot.

	\subsection{Qualitative Multi-environment behaviours}
	
	In Figure \ref{fig3}, we provide examples of the behaviours of the agent in different environments.
	The agent has different behaviour when dropped to a random position in each environment.
	The vector fields are formed by examining the trajectories from all possible starting positions to the target, and the green trajectory shows one complete example trajectory.
	In open spaces such as the simulated living room \ref{exp2_2} and the simulated kitchen \ref{exp2_1}, the agent tends to make strides and turns for localization. In narrow spaces such as the simulated bathroom \ref{exp2_3} the agent would prioritize moving away from walls. In the real corridor \ref{exp2_7} the agent would have much less turning actions during the narrow hallway but shows turning behaviours in the middle section where the space is relatively open.
	
	\subsection{Generalization and Noise Resilience}
	\begin{figure}[h]
		\centering
		\vspace{-3mm}
		\includegraphics[width=0.28\textwidth]{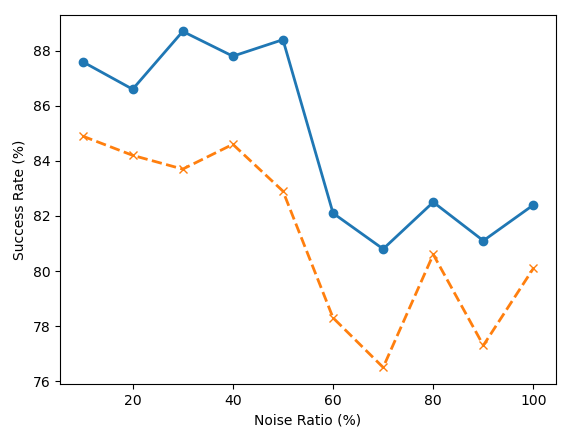}
		\caption[Agent success rate verses noise ratio]{Success Rate drops with increasing noise ratio in current view sampling. The blue line indicates \ac{merlin}'s performance, the orange line indicates the joint expert.}
		\vspace{-5mm}
		\label{fig4}
		\vspace{-2mm}
	\end{figure}
	In the third experiment, the time limit is tightened for completing the episode, and the noise ratio for current view sampling is increased.
	The sampling method is also changed from Gaussian noise to a uniform noise scaling with the grid size to increase difficulty.
	A simple size of 50 is used for each noise level, and the agent will perform 1000 episodes across the 4 simulated environments.
	As shown in Figure \ref{fig4}, \ac{merlin} outperforms the joint expert consistently and maintains more than 80\% success rate until the noise level reaches 100\% of the grid size.
	A drop in performance occurs around 50\% noise level. 
	This is due to the possible sampling positions of each grid starting to overlap with each other.
	At 100\% noise level, the sampling position can drift to another state's position.
	These results indicate the \ac{merlin} agent has a good generalization ability within each environment and is not over-fitting to the training environment.
	This also indicates that an agent trained off-line in a gridded version of a real-world dataset could potentially be transferred to operate in the real world.

	\subsection{Live Demonstration}
	\vspace{-0.7mm}
	In the last experiment, the live demo shows the Turtlebot performs a target-driven navigation task in the real-world location of Env7 (real corridor) using the \ac{merlin} model (\href{https://youtu.be/ayjwNdCOkbw}{Link to video}). 
	The robot successfully completed the task with only a few missteps outside the optimal path, likely caused by lagging and drifting errors.
	We can also observe the robot making a straight line in the hallway, but turning and making strides in the relatively open area in the middle.
	Compared to the gridded simulated environments, the robot spends more time turning for localization and attempted correction to drifting errors. 
	As the model is trained in discrete environments, it has a natural deficiency in handling inaccurate turning angles, this could potentially be improved by generalizing over rotation during training.
	The model's solution to this is to keep turning until it finds a recognizable direction.
	However, this strategy has difficulties under excessive lagging and drifting error.
%
	
	\section{CONCLUSIONS}
	
	We have introduced the multi-environment navigation problem in the field of robotic navigation and proposed a multi-task deep reinforcement learning framework to approach this problem through visual navigation. 
	Overall, the \ac{merlin} model outperforms the \ac{sota} model in the multi-environment target-driven navigation tasks in both performance and training speed.
	Interestingly, \ac{merlin} also outperforms specialist single-environment expert networks even on their own training environment.
	We observed different behaviours depending on the surroundings in both the simulated environments and real-world environments.
	We also demonstrated the model's ability in operating in the real world even when trained off-line in discrete environments.
	
	It is foreseeable that upcoming robots will require more multi-tasking capability than navigation in multiple different environments. 
	They may also need adaptive skills for undertaking various non-navigation tasks in a variety of locations.
	Mimicking humans ability to adapt to environments is going to be vital for robots and provides a great challenge for the field of robotics and artificial intelligence. 
	As for future work, it may prove useful to focus on the robot's capability to transition smoothly between expert networks over time. 
	Another important topic is allowing the robot to adjust its behaviour to adapt to different network resource allocation, in relevance to lifelong learning. 
	As well as the robot's ability of understanding a complex task and break them down to multiple simpler, more manageable tasks each with an independent expert network.

\section*{ACKNOWLEDGMENT}
	
	This work was partially supported by the UK Engineering and Physical Sciences Research Council (EPSRC) grant agreement EP/S035761/1 and Innovate UK Autonomous Valet Parking Project (Grant No 104273).

	\bibliography{myfirstpaper.bib}
	\bibliographystyle{plain}

\end{document}